\newif\ifarxiv
\arxivtrue

\documentclass{article}
\usepackage{float}

\usepackage{arxiv}

\usepackage{iftex}
\ifPDFTeX
  \usepackage[utf8]{inputenc} 
  \usepackage[T1]{fontenc}    
\else
  \usepackage{fontspec}       
\fi
\usepackage{hyperref}       
\newcommand{\mainref}[1]{\ref{#1}} 
\usepackage{url}            
\usepackage{booktabs}       
\usepackage{amsfonts}       
\usepackage{amsmath}

\DeclareMathOperator*{\argmax}{\arg\!\max  \,}
\usepackage{nicefrac}       
\usepackage[nopatch=footnote]{microtype} 
\usepackage{lipsum}		
\usepackage{graphicx} 
\usepackage{doi}
\usepackage{xcolor}
\usepackage[normalem]{ulem} 
\usepackage{caption}
\usepackage{amsfonts}
\usepackage{dsfont}
\usepackage{comment}
\usepackage{bbm}
\usepackage{setspace}
\ifarxiv\else
\fi
\usepackage[mathlines]{lineno}
\ifarxiv\else
  \linenumbers 
\fi

\usepackage[
  backend=biber,
  style=authoryear,
  dashed=false,
  natbib=true,
  giveninits=true,
  uniquename=false,
  uniquelist=false,
  date=year,
  url=false,
  doi=false,
  isbn=false,
  eprint=false,
  maxbibnames=3,
  minbibnames=3, 
  maxcitenames=2
]{biblatex}

\renewbibmacro{in:}{%
  \ifentrytype{article}{}{\printtext{\bibstring{in}\intitlepunct}}}

\AtEveryBibitem{%
  \clearfield{eprint}%
  \clearfield{eprinttype}%
  \clearfield{eprintclass}%
  \clearfield{note}%
  \iffieldundef{booktitle}{}{\clearfield{eventtitle}}
  \clearfield{number}
  \clearfield{month}%
  \clearfield{version}%
  \clearfield{pubstate}
  \clearfield{howpublished}%
  \clearfield{venue}%
  \clearfield{series}%
  \clearlist{publisher}%
  \clearlist{location}%
  \clearlist{organization}%
  \clearfield{doi}%
  \clearfield{url}
  \clearfield{urlyear}
  \clearfield{urlmonth}%
  \clearfield{urlday}%
  \clearfield{urlendyear}%
  \clearfield{urlendmonth}%
  \clearfield{urlendday}%
}

\title{Automated open-set multi-view seahorse re-identification from underwater photograph sequences}

\ifarxiv
\author{ \hspace{1mm}Tristan Fauvel \thanks{Corresponding author. tristan.fauvel@baysic-labs.com} \\
	Baysic Labs \\
    37 rue Sommeiller, 74000 Annecy, France
    \And 
	Sylvie Louisy\\
	Association Peau-Bleue \\
    46 Rue des Escais, 34300 Agde, France
    \And 
	Patrick Louisy \\
	Association Peau-Bleue \\
    46 Rue des Escais, 34300 Agde, France \\
    UMR BOREA (MNHN, CNRS, Sorbonne Université, IRD, UniCaen, UGA), \\ Station marine de Dinard, CRESCO \\
    38 rue du Port Blanc, 35800 Dinard, France
	\And 
	Thomas Fauvel \\
    Parc Naturel Marin du Bassin d'Arcachon, \\
    4 Rue Copernic, 33470 Le Teich, France \\
    Office Français de la Biodiversité, \\
    12 Cours Lumière, 94300 Vincennes, France  
}
\else
\author{}
\fi

\renewcommand{\undertitle}{} 
\renewcommand{\headeright}{} 
\renewcommand{\shorttitle}{Multi-view open-set seahorse re-identification} 

\hypersetup{
hidelinks, 
pdftitle={Automated open-set multi-view seahorse re-identification from underwater photograph sequences},
pdfsubject={cs.CV, q-bio.QM, q-bio.PE},
pdfauthor={\ifarxiv Tristan Fauvel, Sylvie Louisy, Patrick Louisy, Thomas Fauvel\fi},
pdfkeywords={animal re-identification, capture-mark-recapture, Hippocampus, local feature matching, open-set recognition, photo-identification, seahorses, sex determination},
}

\begin{document}
\maketitle

\ifarxiv
\begin{abstract} 
    Automated animal re-identification from photograph sequences can facilitate minimally invasive Capture-Mark-Recapture studies at scale. Yet, photographs taken during an observation session can vary greatly in quality, not always providing the necessary information for accurate re-identification. 
    In this paper, we present new methods for open-set re-identification from image sequences. We train and evaluate taxonomic classification, sex determination and re-identification models on a dataset of 35,853 photographs of 473 individual seahorses observed in the wild.
    Combining the photographs of an observation improves all three tasks. For species and sex, a linear probe on DINOv3 embeddings classifies each photograph, and we average the logits over the photographs of an observation. 
    For open-set re-identification, we retrieve candidate individuals with a fine-tuned MiewID descriptor, and weight each photograph by a learned relevance score. Local feature matching with ALIKED and LightGlue combined with a geometric check with RANSAC that keeps only the consistent matches then reranks the candidates. A logistic model estimates the probability that the individual is already in the gallery. A maximum a posteriori rule then chooses between the best-matching identity and a new individual. This method reaches a normalised accuracy of 0.91 when the gallery holds every past observation of each individual, and 0.82 when it holds only one. 
\end{abstract}
\else
\begin{abstract} 
    1. Automated animal re-identification from photograph sequences can facilitate minimally invasive Capture-Mark-Recapture studies at scale. Yet, photographs taken during an observation session can vary greatly in quality, not always providing the necessary information for accurate re-identification.

    2. In this paper, we present new methods for open-set re-identification from image sequences. We train and evaluate taxonomic classification, sex determination and re-identification models on a dataset of 35,853 photographs of 473 individual seahorses observed in the wild.

    3. Combining the photographs of an observation improves all three tasks. For species and sex, a linear probe on DINOv3 embeddings classifies each photograph, and we average the logits over the photographs of an observation.

    4. For open-set re-identification, we retrieve candidate individuals with a fine-tuned MiewID descriptor, and weight each photograph by a learned relevance score. Local feature matching with ALIKED and LightGlue combined with a geometric check with RANSAC that keeps only the consistent matches then reranks the candidates. A logistic model estimates the probability that the individual is already in the gallery. A maximum a posteriori rule then chooses between the best-matching identity and a new individual. This method reaches a normalised accuracy of 0.91 when the gallery holds every past observation of each individual, and 0.82 when it holds only one. 
\end{abstract}
\fi


\section{Introduction}
  
\subsection{Animal re-identification from photographs}
Animal re-identification (or re-ID) involves assigning an identity to an individual based on distinctive characteristics such as body patterns or other morphological traits.
Animal re-identification from photographs taken in the wild has emerged as a powerful non-invasive Capture-Mark-Recapture (CMR) tool with several practical advantages over artificial markers \parencite{glennie_estimating_2021}, including reduced costs, reduced harm or stress, potential for public participation in photo-based CMR programmes \parencite{gould_dragon_2024}, and large-scale use of camera traps \parencite{schneider_past_2019, alexander_novel_2022}.
However, manual photo cataloguing by experts is time-consuming and hinders its large-scale deployment \parencite{connor_individual_2001, irvine_evaluation_1982}.
Therefore, automated animal re-identification through computer vision has become a promising solution \parencite{bolger_computerassisted_2012, schneider_past_2019, reno_siftbased_2019}. 
In particular, recent advances in deep learning have shown that models can achieve high performance in re-identifying individuals across various species \parencite{vidal_perspectives_2021, beyan_speciesspecific_2026}, sometimes exceeding human observer capability \parencite{schneider_similarity_2020}.
In practice, during a single encounter with an animal, an observer or a camera trap will typically capture a sequence of photographs, which may provide different viewing angles, distances, or lighting, and together can provide more information about the animal's identity than photographs taken in isolation.
In this work, we develop methods for automated taxonomic classification, sex determination, and animal re-identification from camera sequences, leveraging the complementary information from multiple photographs while accounting for their variable informativeness relative to the animal characteristics and identity. We apply this approach to an unpublished dataset of expert-reidentified sequences of underwater photographs of wild seahorses from two species, the long-snouted seahorse (Hippocampus guttulatus) and the short-snouted seahorse (Hippocampus hippocampus) from the Thau lagoon, on the French Mediterranean coast.
 
\subsection{Re-identification methods for seahorses}

Seahorse populations are experiencing a marked decline worldwide, which may be linked to fishing pressure due to illegal trade, habitat loss and warming waters \parencite{harasti_declining_2016, pierri_density_2021, correia_monitoring_2022, vivas_population_2024}.
Specifically, Mediterranean \textit{H. guttulatus} and \textit{H. hippocampus} populations showed sharp density declines \parencite{caldwell_revisiting_2012, pierri_density_2021, correia_monitoring_2022, vivas_population_2024}.
However, the population dynamics of many seahorse species remain poorly understood.
Project Seahorse highlights that 17 seahorse species are data-deficient, so their conservation status could not be assessed \parencite{pollom_global_2021}.
Reliably estimating seahorse demographic parameters remains difficult due to the limitations of traditional CMR or transect methods which are complex to implement for these cryptic, low-density fish with camouflage behaviour \parencite{curtis_life_2006}.  
The most common artificial tagging method in seahorses is the application of visible implant fluorescent elastomer (VIFE) tags, which involves injecting seahorses with long-lasting colourants, allowing for individual recognition upon recapture \parencite{woods_visible_2004, monteiro_homing_2005, curtis_visible_2006, sogabe_mate_2007, harasti_getting_2021}.
However, they require trained professionals and do not readily scale to citizen-science programmes.
\textit{H. guttulatus} and \textit{H. hippocampus} can be distinguished based on head, snout, and trunk shape \parencite{lourie_guide_2004}.
\textit{H. guttulatus} has a distinct mark pattern composed of white spots scattered throughout the entire body surface, allowing for photo-identification by human experts with manual labelling of spots and image comparisons \parencite{correia_use_2014}.
Moreover, \textcite{correia_crown_2021} demonstrated that it is possible to re-identify \textit{H. hippocampus} from photographs taken in captivity based on crown characteristics.
However, in the wild, the presence of fouling on the seahorse body can make it difficult to distinguish these patterns.

\subsection{Contributions}

In this paper, we present the following contributions.
First, we develop multi-view seahorse taxonomic classification and sex determination models that leverage multiple photographs of a single seahorse observation to outperform approaches based on single images. 
Second, we present a new method for animal re-identification that: 
\begin{itemize}
    \item Uses a Bayes-optimal decision rule for open-set re-identification.
    \item Combines a fine-tuned foundation model trained on large multi-species corpora for the extraction of global descriptors, coupled with a local pattern matching module.  
    \item Leverages an arbitrary number of photographs to enable reliable animal re-identification, weighting their relevance based on estimated image informativeness. 
\end{itemize}
We demonstrate that this approach yields high performance in seahorse re-identification.

\subsection{Related work}

Modern photo re-ID techniques based on deep learning can be broadly categorised into two approaches. The first one relies on deep neural networks trained to produce image embeddings (also called global embeddings); the second relies on the extraction and matching of local features. The global features approach is typically framed as a metric learning problem. A deep neural network is trained to minimise a triplet loss or an ArcFace loss \parencite{deng_arcface_2019}. At inference time, the cosine similarity between the embeddings of each query image and gallery image is computed, and a similarity threshold is used to decide whether the image belongs to an individual already observed or not. In the former case, a ranked list of candidate matches is produced. Recently, foundation models trained on many datasets from different species have shown strong performance in re-identification, both in zero-shot learning and after fine-tuning on a specific dataset \parencite{cermak_wildlifedatasets_2024, otarashvili_multispecies_2024}. Local features are extracted by deep neural networks such as ALIKED \parencite{zhao_aliked_2023}, and are matched using a deep neural network specialised for feature matching, such as LightGlue \parencite{lindenberger_lightglue_2023}. This approach is usually more accurate than global feature matching, but since it performs pairwise comparison between the query and the gallery images, it becomes computationally impractical as the dataset grows. In order to get the most out of the slow but accurate local features approach and the fast but less accurate global features, \textcite{cermak_wildfusion_2024} introduced WildFusion, which uses global feature similarity to create a list of the most likely matching candidates from the gallery, on which local feature matching is applied. The scores are then calibrated using isotonic regression and combined using a weighted average.

\textcite{spoto_taxonomic_2024} developed two models for taxonomic classification of \textit{H. hippocampus} and \textit{H. guttulatus} on a dataset of seahorse photographs from the Mar Piccolo of Taranto. 
They found that a VGG16 architecture achieved an AUPRC (Area Under the Precision-Recall Curve) of about 0.97. However, despite the dataset containing multiple images of the same individuals, sometimes extracted from a video which likely creates near-duplicate images of the same individual, the authors do not mention any data leakage mitigation strategy. 

\textcite{aloui_individual_2022} reported promising preliminary results in the automated re-identification of \textit{H. guttulatus} using a dataset related to ours.
They trained EfficientNet variants \parencite{tan_efficientnet_2020} using triplet loss \parencite{schroff_facenet_2015}.
However, they only evaluated their approach in a closed-set setting, where the query images correspond to individuals always present in the gallery.
In this setting, they reported top-1 accuracy of up to 76\% on manually curated images where the head is clearly visible.

\textcite{bergamini_multiviews_2018} proposed a multi-view method for cattle re-identification. In this method, two different networks receive a front and a side image of the head.
The resulting embeddings are then combined and fed to another network that fuses the representations.
However, this hybrid fusion approach does not accommodate more than two images, and requires manual annotation of front and side images.

Set-to-set recognition methods from the face and person re-identification literature also address the many-to-many scenario, when a query is made of multiple views of the same individual with varying relevance to the re-identification task. The Quality Aware Network (QAN) learns a quality score for each image, and pools the image features by their quality score \parencite{liu_quality_2017}.
Probabilistic Face Embeddings (PFE) represent each image as a multivariate Gaussian distribution in a latent space \parencite{shi_probabilistic_2019}.
The mean of the distribution is the estimate of the most likely feature values while the variance shows the uncertainty in the feature values. Embeddings of a set are fused by inverse-variance weighting, and two sets are compared by the mutual likelihood score. Both methods learn the image weights without labels on image quality.

\section{Materials and methods}

\subsection{Data}
Photographs of \textit{H. hippocampus} and \textit{H. guttulatus} were taken underwater by marine biologists between March 2022 and October 2025, over 150 survey night dives in the Thau lagoon.
Species, sex and identity were determined by one of the authors.
The dataset contains photographs of 460 \textit{H. guttulatus} and 13 \textit{H. hippocampus} individuals (see Table \ref{tab:summary_statistics_by_species_and_sex}).
Among the individuals seen on at least two dates (272 \textit{H. guttulatus} and 8 \textit{H. hippocampus}), the median time between consecutive observations of the same individual was 13 days and 7 days respectively (Table \ref{tab:inter_observation_gap_statistics_by_species}). 
 
The data were not specifically collected for training machine learning models, and therefore reflect standard field experiment conditions, where the seahorse photographs were taken from different distances, viewing angles, lighting conditions, and backgrounds.
The photographers tried to capture both fine-grained details of the head and the overall environment.
Some photographs also contain juvenile individuals.
We did not manually curate the data to better capture the performance of the models in a field deployment setting, in contexts such as citizen science programmes.
Moreover, the extreme class imbalance (98\% of observations correspond to \textit{H. guttulatus}) reflects natural species observation imbalance.

\begin{figure}[ht]
    \centering
    \includegraphics[width=1\textwidth]{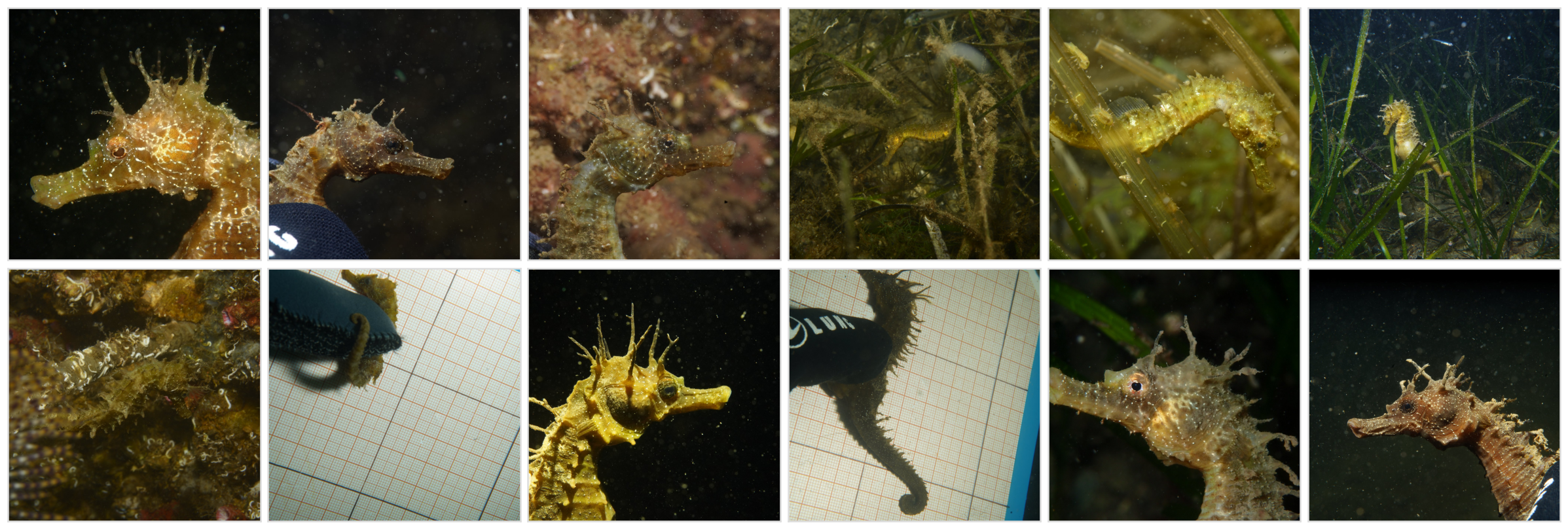}
    \caption{Illustrative examples of photographs from the dataset.
    As this dataset was not specifically collected for the purpose of automated re-identification, images vary greatly in quality.
    Some images focus on the seahorse's head, while others show the individual in its environment, with potential occlusion.
    Some images were taken for the purpose of measuring the animal's length.}
    \label{fig:fig1_diversity}
\end{figure}

\begin{table}[htbp]
\centering
\footnotesize
\setlength{\tabcolsep}{5pt}
\begin{tabular}{llrrrr}
\toprule
Species & Sex & N & FU med/max (d) & Obs/ind. (IQR) & Photos/obs (IQR) \\
\midrule
\textit{H. guttulatus} & Male & 212 & 40/1015 & 3 (1--7) & 13 (10--17) \\
\textit{H. guttulatus} & Female & 185 & 37/1094 & 2 (1--8) & 12 (10--16) \\
\textit{H. guttulatus} & Unknown & 63 & 0/41 & 1 (1--1) & 15 (12--19) \\
\textit{H. hippocampus} & Male & 8 & 2/87 & 2 (1--2) & 16 (13--39) \\
\textit{H. hippocampus} & Female & 5 & 39/108 & 5 (5--7) & 13 (10--32) \\
\bottomrule
\end{tabular}
\caption{Summary statistics by species and sex.
N: number of individuals.
FU: follow-up time in days.
Obs/ind.: number of observations per individual.
Photos/obs: number of photos per observation.
IQR: interquartile range.}
\label{tab:summary_statistics_by_species_and_sex}
\end{table}

\begin{table}[htbp]
\centering
\begin{tabular}{lrrrrrrr}
\toprule
Species & n gaps & Min (d) & P25 (d) & Median (d) & P75 (d) & Max (d) & Mean $\pm$ SD (d) \\
\midrule
\textit{H. guttulatus} & 1981 & 1 & 7 & 13 & 31 & 418 & 33 $\pm$ 57 \\
\textit{H. hippocampus} & 36 & 1 & 1 & 7 & 14 & 77 & 12 $\pm$ 17 \\
\bottomrule
\end{tabular}
\caption{Inter-observation gap statistics by species.
SD: standard deviation.
P25: 25th percentile.
P75: 75th percentile.}
\label{tab:inter_observation_gap_statistics_by_species}
\end{table}

\subsection{Fieldwork ethics}
We used existing photographs from a past expert photo-identification project in the Thau lagoon, France. This study did not involve animal handling. During this previous project, divers held some seahorses by hand for a few seconds. Animal ethics committee approval was not required for this field protocol under French regulations.

\subsection{Image preprocessing}
Seahorses were detected and segmented in images using the Segment Anything Model 3 (SAM 3) \parencite{carion_sam_2025}, which requires no manual annotation of the data (prompt: "seahorse", threshold value of 0.7 for detection, and 0.6 for the segmentation mask confidence).
Crops corresponded to detection boxes with a 5\% padding.
The segmentation masks were used to set the background to white.
The cropped and segmented images were white-padded to be square and resized to the models' input dimensions.
A photograph can show more than one seahorse, and the seahorse body can be obstructed, resulting in more than one detection per image. For each image, we kept the crop with the highest SAM 3 detection confidence.
SAM 3 found at least one seahorse in 30,439 photographs, leaving five observations without detection. 

\subsection{Data splits for species and sex classification}

To evaluate taxonomic and sex classification odels, we used stratified 5-fold cross-validation, with all images of an individual assigned to the same fold. We stratified the folds by class, balancing class proportions across folds by individual count. Evaluation metrics were computed on pooled out-of-fold
predictions.

Within each fold, we chose the hyperparameter (regularisation strength $C$ of the logistic probe) by a nested 5-fold cross-validation over the training individuals only, stratified by class. We kept the $C$ value with the best identity-balanced accuracy on the pooled inner validation fold predictions. We then refitted the probe on the full training fold with this $C$ value.

\subsection{Taxonomic and sex classification models}
\label{sec:classification_models}

Let the class label be $Y$, and let an observation be a set of photographs: $X_{1:m} = \left(X_1, \cdots, X_m \right)$.
The binary cross-entropy loss is minimised by the posterior predictive class probability $P(Y=1| X_{1:m}, \mathcal{D})$, where $\mathcal{D}$ is the labelled dataset. 
A typical approach would be to aggregate the image embeddings using a learned fusion rule that can accommodate a variable number of views $m$.
Given the extreme species imbalance in our dataset and the limited effective sample size, we focused on linear probing of vision foundation models using a score fusion approach, which first maps each image embedding to a score: $\ell_i = f(X_i)$, then combines the scores into an observation-level posterior distribution of the class label $P(Y|\ell_{1:m})$, with $\ell_{1:m} = (\ell_1, \dots, \ell_m)$.

Let $L$ be the posterior log odds (or logit) of the observation, $L = \log \frac{P(Y=1|X_{1:m}, \mathcal{D})}{P(Y=0|X_{1:m}, \mathcal{D})}$. The logit-sum fusion rule corresponds to $P(Y=1|\ell_{1:m})=\sigma(L)=\sigma\left(\sum_{i=1}^{m}\ell_{i}\right)$, whereas the average-logit fusion rule is $P(Y=1|\ell_{1:m}, \mathcal{D}) = \sigma \left(\frac{1}{m}\sum_{i=1}^m \ell_i \right)$. The rationale behind these fusion rules is detailed in Appendix \ref{sec:late_fusion_rules}.

In our experiments, we evaluated image-level classification performance, average-logit fusion, and logit-sum fusion. Additionally, we also considered majority vote, in which we assign each image in an observation to its predicted class, and take the fraction of images in each class as the class probability. For sex classification, we excluded the 63 individuals of unknown sex.

For image-level taxonomic and sex classification, we trained logistic regression models on the embeddings obtained from two different vision foundation models: either BioCLIP 2.5 \parencite{gu_bioclip_2025} or DINOv3 ViT-L/16 CLS \parencite{simeoni_dinov3_2025}. Images fed to BioCLIP 2.5 were normalised using OpenAI CLIP statistics, whereas images fed to DINOv3 were normalised using ImageNet statistics. For both models, images were resized to 224×224 pixels. We applied L2 regularisation and tuned the inverse regularisation strength coefficient using a grid search over 49 log-spaced values in $[10^{-4}, 10^{2}]$.


\subsection{Re-identification model}

\subsubsection{Bayes-optimal decision rule}
\label{sec:bayes_rule}

Let $\mathcal{G}$ denote the set of identities in the gallery.
Let $\mathcal{Q}$ denote the set of individuals in the query.
For individual $q \in \mathcal{Q}$, $\mathcal{O}_q$ is the set of query observations of this individual.

Animal re-identification corresponds to determining, for a query observation composed of a sequence of images $X_{1:m}$ of individual $q\in \mathcal{Q}$, if $q \in \mathcal{G}$ (i.e. $q$ belongs to the gallery), and if so, to match $q$ to an individual in the gallery.
Animal re-identification is therefore an open-set recognition task, in the sense that the individuals in the queries at inference time are not necessarily present in the training set, and in the sense that they are not necessarily present in the gallery either. 

The Bayes-optimal decision rule is to return, for a query observation $X_{1:m}$ and given training data $\mathcal{D}$, a predicted identity $\hat{q}$ defined as: 
\begin{equation}
 \begin{aligned}
 \hat{q} &= \underset{q \in \mathcal{G} \cup \{\text{new}\} }{\operatorname{argmax}}P(Y=q|X_{1:m}, \mathcal{D}),
\end{aligned}
\label{eq:bayes_optimal_rule}
 \end{equation}
where $\text{new}$ corresponds to the case where it is predicted that the corresponding individual does not belong to the gallery.

For each observation, we need to estimate two quantities: the probability that the observation corresponds to a known individual, $p_\text{known}=P(K=1|X_{1:m}, \mathcal{D})$, and the closed-set re-identification probability $s_q = P(Y=q|X_{1:m}, K=1, \mathcal{D})$. Let $\hat{z} = \operatorname{argmax}_{q \in \mathcal{G}} s_q$ be the closed-set re-identification decision, and $s_{\hat{z}}$ its probability. The decision is given by (see Appendix \ref{app:open-set_decision_rule} for derivation):

\begin{equation}
\begin{aligned}\hat{q} & =\begin{cases}\hat{z}, & \text{if } \frac{1}{s_{\hat{z}}+1}<p_{\text{known}},\\ \text{new}, & \text{otherwise}.\end{cases}\end{aligned}
\label{eq:map_decision_rule}
\end{equation}
We write $\hat{\gamma} = \mathds{1}\left\{\frac{1}{s_{\hat{z}}+1} < p_\text{known}\right\}$ for the knownness decision, so $\hat{q} = \hat{z}$ if $\hat{\gamma} = 1$, and $\hat{q} = \text{new}$ otherwise.
For query observation $i$, we write these decisions $\hat{z}_i$, $\hat{\gamma}_i$ and $\hat{q}_i$.
 
\subsubsection{Evaluation metrics}
\label{sec:reid_metrics}

Standard top-k accuracy can be misleading due to the fact that some individuals are over-represented among observations (see Table \ref{tab:summary_statistics_by_species_and_sex}).
Therefore, \textcite{adam_wildlifereid10k_2025} recommend using balanced accuracy metrics to evaluate performance \parencite{mendesjunior_nearest_2017}: Balanced Accuracy on Unknown Samples (BAUS), Balanced Accuracy on Known Samples (BAKS), and their geometric mean, the Normalised Accuracy (NA) (detailed in Appendix \ref{app:reid_metrics}).

Additionally, we define the Balanced Re-identification Accuracy as: 
\begin{equation}
\begin{aligned}
    \mathrm{BRA} &= \frac{1}{\left|\mathcal{Q}\right|}\sum_{q \in \mathcal{Q}} \frac{1}{\left|\mathcal{O}_{q}\right|} \sum_{i \in \mathcal{O}_{q}} \left[\mathds{1}\left(\hat{z}_i=q , \hat{\gamma}_i = 1, q \in \mathcal{G} \right) + \mathds{1}\left(\hat{\gamma}_i=0, q \notin \mathcal{G} \right) \right] \\
      &= \frac{\rho }{\left|\mathcal{G}\cap \mathcal{Q}\right|}\sum_{q \in \mathcal{G} \cap \mathcal{Q}} \frac{1}{\left|\mathcal{O}_{q}\right|} \sum_{i \in \mathcal{O}_{q}} \mathds{1}\left(\hat{q}_i=q \right) + \frac{1-\rho}{\left|\bar{\mathcal{G}} \cap \mathcal{Q}\right|}\sum_{q \in \bar{\mathcal{G}}\cap \mathcal{Q}} \frac{1}{\left|\mathcal{O}_{q}\right|} \sum_{i \in \mathcal{O}_{q}} \mathds{1}\left(\hat{q}_i=\text{new} \right) \\
      &= \rho\mathrm{BAKS} + (1-\rho)\mathrm{BAUS}
\end{aligned} 
\label{eq:bra}
\end{equation}
where $\rho = \frac{|\mathcal{G} \cap \mathcal{Q}|}{|\mathcal{G} \cap \mathcal{Q}| + |\bar{\mathcal{G}} \cap \mathcal{Q}|}$ is the share of known-query individuals. 
 
The known vs unknown individual decision depends on a tuned threshold applied to decision score (except for the maximum a posteriori (MAP) open-set decision rule, see section \ref{MAP}). Each threshold gives one pair (BAKS, BAUS). The area under the BAKS vs BAUS curve is a balanced version of the open-set classification rate \parencite{dhamija_reducing_2018}, which measures the trade-off between accuracy and open-set detection rate as a threshold on the confidence of the predicted class is varied (denoted AUOSCR).  
 
\subsubsection{Data splits for re-identification}
\label{sec:reid_splits}
For re-identification, given the computational cost of CNN fine-tuning, we used a single identity-disjoint split (70\% train / 10\% val / 20\% test).
Identities were stratified by their number of observations so that each split receives a comparable mix of sparse and well-sampled individuals.

Within the validation and test sets, we used each observation in turn as a query. To make an unknown query, we hid every observation of the query's individual in the gallery. To make a known query, we hid only the query observation, so that the gallery still holds the individual's other observations. A known query is therefore possible only for an individual with at least two observations.

\subsubsection{Gallery masking policy for evaluation}
We considered two gallery policies.
Under the "rich gallery" policy, the gallery holds every observation of the set except those hidden for the query (Section~\ref{sec:reid_splits}).
Under the "sparse gallery" policy, each individual in the gallery contributes a single observation, which simulates a scenario with less data.
We drew this observation uniformly at random among the individual's remaining observations, independently for each query.
We repeated the sparse-gallery evaluation with five independent draws. For each draw, we refitted every threshold and model on the validation split and applied them to the test split.

\subsubsection{Modelling approaches}
\label{sec:modelling_approaches}
\paragraph{MiewID with zero-shot learning }

We first evaluated the performance of MiewID in a zero-shot setting, that is, without fine-tuning on our seahorse dataset (step 1).
For a query observation $X_{1:m}$ and a gallery individual $g$ with images $X^{(g)}_{1:n}$, we computed the cosine similarity between the image embeddings $S_C(X_i, X^{(g)}_j)$ for any query image/gallery image pair.
For each query image, we then applied max-pooling on the gallery images, resulting in a vector with entries defined as $\underset{j}{\max} \; S_C(X_i, X^{(g)}_j)$.
This corresponds to selecting, for each query image/gallery identity pair, the gallery image most similar to the query image. 

After this gallery-side pooling step, similarity scores were averaged over query images, resulting in query observation/gallery identity similarity scores:
\begin{equation}
\label{eq:image_aggregation_baseline}
S(X_{1:m},g) = \frac{1}{m} \sum_{i=1}^m \underset{j}{\max \,} S_C(X_i, X^{(g)}_j)
\end{equation}

A common approach to open-set animal re-identification is to define a decision threshold $\tau$. A query with a score above $\tau$ is assigned to an existing individual in the gallery (the top match), otherwise it is considered a new individual \parencite{liao_openset_2014, pakhomov_notebook_2025, kabuga_similarity_2024, adam_wildlifereid10k_2025}.
That is: 
 
\begin{equation}
\hat{q} =
\begin{cases}
\underset{g \in \mathcal{G}}{\argmax } P(Y = g | X_{1:m}, \mathcal{D}), & \text{if } \underset{g \in \mathcal{G}}{\max \,} P(Y = g | X_{1:m}, \mathcal{D}) \geq \tau, \\
\text{new}, & \text{otherwise}.
\end{cases}
\label{eq:known_unknown_decision_heuristic}
\end{equation}

We evaluated this approach by applying the rule: 
 
\begin{equation}
\hat{q} =
\begin{cases}
\underset{g \in \mathcal{G}}{\argmax} \; S(X_{1:m},g), & \text{if } \underset{g \in \mathcal{G}}{\max} S(X_{1:m},g) \geq \tau, \\
\text{new}, & \text{otherwise}.
\end{cases}
\end{equation}

We calibrated $\tau$ by maximising normalised accuracy on the validation split using grid search. The grid was built from 51 evenly spaced quantiles of the score distribution in the validation set. 

\paragraph{MiewID fine-tuning}

MiewID was trained on a wide diversity of species, but not on seahorses \parencite{otarashvili_multispecies_2024}. We therefore fine-tuned the pretrained MiewID-msv3 backbone (2152-dimensional embeddings, $440\times440$ input) on our dataset following the method described in \textcite{yesharim_nearperfect_2026}, detailed in Appendix \ref{app:miewid_finetuning} (step 2 of the pipeline).

\paragraph{Relevance weighting}

Query image quality varies greatly within an observation (Figure~\ref{fig:fig1_diversity}). 
We therefore explored the use of a new approach to weight the contribution from each query image based on its relevance to re-identification, by generalising the approach described in equation \ref{eq:image_aggregation_baseline}.
Observation-level similarity scores were computed by assigning query-image weights: 
\begin{equation}
\label{eq:image_aggregation_weighted}
S(X_{1:m},g) = \sum_{i=1}^m w(X_i) \underset{j}{\max \;} S_C(X_i, X^{(g)}_j),
\end{equation}
with $\sum_{i=1}^m w(X_i) = 1$.

We hypothesised that DINOv3 embeddings contained information regarding the usefulness of a given image for re-identification. Therefore, we divided the observations in the validation split into a fitting fraction (65\% of the observations) and a scoring fraction (35\%), and trained an L2-regularised logistic regression classifier on the fitting fraction of the validation split to predict whether image-level closed-set rank-1 re-identification from the fine-tuned MiewID was correct or not, with standardised DINOv3 CLS embeddings as input. We selected the inverse regularisation strength $C$ from 13 log-spaced values between $10^{-4}$ and $10^{2}$, by maximising the AUROC in a 5-fold cross-validation grouped by individual. We called the output probabilities of this model image relevance scores.

%
We then weighted the MiewID-derived similarities by the predicted image relevance scores, floored at 0.02 so that no image is discarded, then normalised to sum to 1, and computed the weighted average similarity score (step 3 of the pipeline). 
We compared relevance weighting with QAN and PFE, which we applied at the same step on the fine-tuned MiewID embeddings (Appendix~\ref{sec:supp_methods}).

For this approach and the ones below, we calibrated $\tau$ by maximising normalised accuracy on the scoring fraction of the validation split using grid search. 

\paragraph{ALIKED/LightGlue reranking}

Inspired by the approach presented in \textcite{yesharim_nearperfect_2026}, we adopted a retrieve-then-rerank strategy in which a shortlist of candidate gallery identities, retrieved with the global fine-tuned MiewID descriptor, is re-ordered for each query observation using local feature matching.

For each query observation, we ranked the gallery identities by the
relevance-weighted similarity score of equation~\ref{eq:image_aggregation_weighted}
and retained the $K = 10$ highest-scoring identities.
For each retained candidate identity, we selected the $L = 10$ query-image/gallery-image pairs
with the highest MiewID cosine similarity, over the photographs of all eligible gallery observations of that identity.
This gives at most $K \times L = 100$ candidate
pairs per query observation.
This choice was made to reduce the computational cost of running local feature extraction and matching. 
Local features were extracted from each image with ALIKED
\parencite{zhao_aliked_2023} (up to 256 keypoints per image, each image resized so that
its longer side does not exceed 512~pixels) and matched with LightGlue
\parencite{lindenberger_lightglue_2023}.
The score of a pair is the number of established correspondences.
Each shortlisted candidate identity was then scored by the mean of its five highest pair scores (step 4 of the pipeline).
This scoring approach was used to avoid reliance on a single spuriously matched pair and dilution by uninformative images.
Within the shortlist, candidates were re-ranked by this pooled score, the MiewID similarity being retained only as a tie-break.

\paragraph{Geometric verification with RANSAC}

\begin{figure}[htbp]
\centering
\includegraphics[width=1\linewidth]{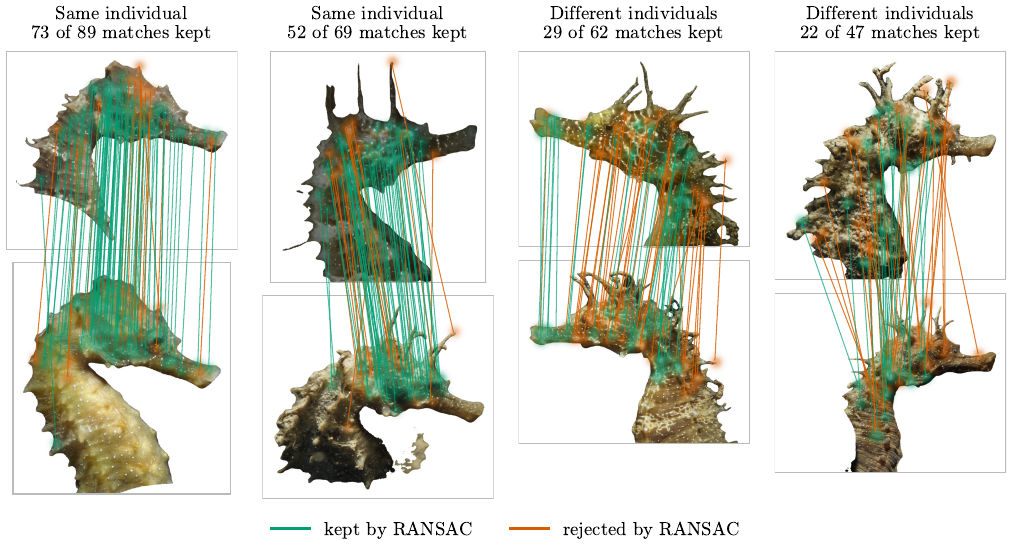}
\caption{ALIKED/LightGlue correspondences for image pairs from the test split, before and after geometric verification.
Green lines join the correspondences that agree with the fundamental matrix fitted by RANSAC, and orange lines join the correspondences that RANSAC rejects as outliers.}
\label{fig:local_matches}
\end{figure}

Some LightGlue correspondences join two points that look alike but are not the same point of the animal. For each image pair, we fitted a fundamental matrix to the correspondences with RANSAC (OpenCV, 2,000 iterations, confidence 0.999, fixed seed). A fundamental matrix relates two views of the same rigid object. A correspondence is an inlier when its point in the second image lies within 4 pixels of the epipolar line that the matrix gives for its point on the first image. 
The score of a pair is then its number of inliers instead of its number of correspondences (step 5 of the pipeline). Figure \ref{fig:local_matches} shows the inliers and outliers of four test pairs. We gave a score of 0 to a pair with fewer than 8 correspondences, the minimum for a fundamental matrix.

\paragraph{Known/unknown probability model}

We investigated an approach for the detection of new individuals, based on explicit modelling of the probability of the query corresponding to a known individual in the gallery. 
Intuitively, the model would benefit from using as features a measure of the match strength between a query and an identity, and a measure of the best match decisiveness.
Therefore, we trained a logistic regression model to predict whether the individual in a query is present in the gallery, taking as input the following features: gap between the best and second-best identity's pooled ALIKED/LightGlue inlier count, fraction of the photographs in an observation whose own image-level MiewID top-1 identity agrees with the observation's best matching identity, mean and standard deviation of the image-level cosine similarity from the query observation's photographs to the best matching identity's photographs determined by local feature matching, and a binary variable corresponding to the agreement between the ALIKED/LightGlue rerank and the fine-tuned MiewID top match (see Appendix \ref{app:known_unknown_model} for details).

We standardised these features and gave each individual the same total weight, so that a heavily photographed individual does not dominate the fit.
We fitted the logistic regression on the fitting fraction of the validation split.
We kept the threshold-based rule of equation \ref{eq:known_unknown_decision_heuristic} but replaced the quantity being thresholded: we used the output of the logistic regression model, $p_\text{known}$, instead of $s_{\hat{z}}$ to determine if the observation corresponds to a new individual or not (step 6 of the pipeline).
We tuned the threshold $\tau$ on the scoring fraction of the validation split, as above.

\paragraph{Maximum a posteriori open-set decision rule}
\label{MAP}
 We used the decision rule of equation \ref{eq:map_decision_rule} as the last step of the pipeline (step 7). It requires two calibrated quantities: the knownness probability $p_\text{known}$ and the closed-set re-ID distribution $P(Y=k | X_{1:m}, K{=}1)$. The procedures to calibrate $p_\text{known}$ and estimate the closed-set matching distribution from pairwise matching probabilities are described in Appendix \ref{app:map_open_set_rule}.

Compared to the threshold-based open-set decision, this rule requires no decision threshold $\tau$. 

\section{Results}
\subsection{Species and sex classification}
\label{sec:classification_results}

Individuals contribute very unequal numbers of images and observations, so all reported classification metrics are identity-balanced.
In order to determine uncertainty related to performance metrics for each sex classification model, we used percentile bootstrap over individuals with 1,000 draws. We give no intervals for species classification due to the small number of \textit{H. hippocampus} individuals that would make the bootstrap intervals unreliable.

Appendix \ref{sec:supp_species} reports the species classification results.  
For species classification, DINOv3 outperformed BioCLIP 2.5, altough this comparison only relies on 13 \textit{H. hippocampus} individuals. 
 We compared the multi-view fusion methods for sex determination (Table \ref{tab:sex_fusion}). 
Leveraging all images from an observation led to higher prediction accuracy than considering single images.
 Balanced accuracy rose from 0.80 to between 0.88 and 0.90.
Since logit-sum and average-logit fusion give the same classification decision, they lead to the same balanced accuracy and macro F1.
 The AUPRC showed the same gain.
With males as the positive class, it rose from 0.90 to between 0.95 and 0.97. A random score would get 0.54. 
 The bootstrap intervals of majority vote overlap with those of the two logit fusion rules.
All three fusion rules lowered the Brier score.
All three rules raised the Maximum Calibration Error (MCE) and the Expected Calibration Error (ECE).
Majority vote raised the ECE for sex the least, from 0.04 to 0.05. Logit-sum fusion gave the highest MCE (0.64), as expected from a rule that treats near-duplicate photographs as independent evidence.

\subsection{Re-identification}

\begin{figure}
     \centering
     \includegraphics[width=1\linewidth]{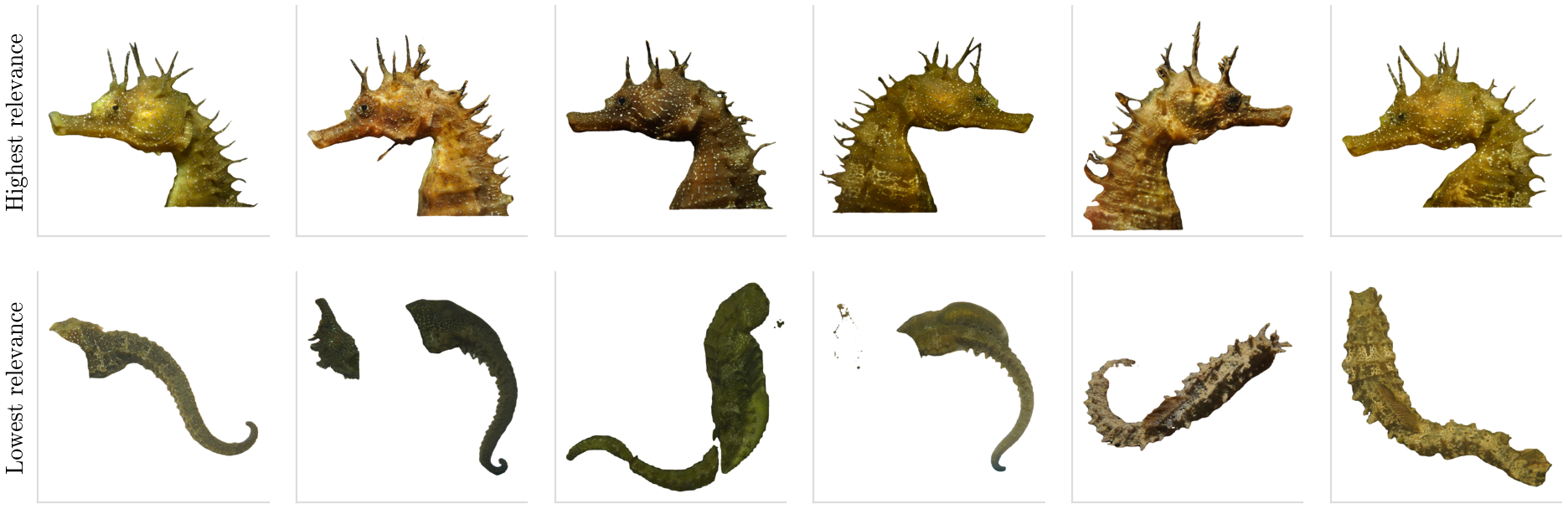}
     \caption{Example of images automatically classified as high relevance  (upper row, usually side view of the head with good illumination) versus low relevance (lower row).}
     \label{fig:relevance_weighting}
 \end{figure}

Tables \ref{tab:reid_progression_one_observation_per_individual} and \ref{tab:reid_progression_all_except_query} report the re-identification performance in the sparse and rich gallery settings, respectively. Combining the photographs of an observation improved re-identification compared to using single photographs (Table \ref{tab:reid_query_granularity_finetuned}). With the fine-tuned MiewID descriptor, observation-level queries raised NA by 0.16 and 0.12 with a sparse and rich gallery, respectively. We observed that MiewID fine-tuning brought substantial closed-set re-identification improvements. With a sparse gallery, relevance weighting did not bring substantial benefits, but with a rich gallery it rose NA by 0.11. Figure~\ref{fig:relevance_weighting} shows that high relevance images are usually side view images of the head with good illumination, whereas low relevance images do not show the profile clearly.
Using QAN or PFE did not bring benefits compared to using relevance weighting in the full system (Table \ref{tab:supp_late_steps}). 

Reranking a shortlist of candidates using ALIKED/LightGlue raised NA by 0.07 with a sparse gallery and by 0.01 with a rich gallery. It raised AUOSCR by 0.11 and  0.05, respectively. 
Geometric verification with RANSAC (step 5) further raised NA and AUOSCR in both gallery settings.

\begin{figure}[htbp]
\centering
\includegraphics[width=1\linewidth]{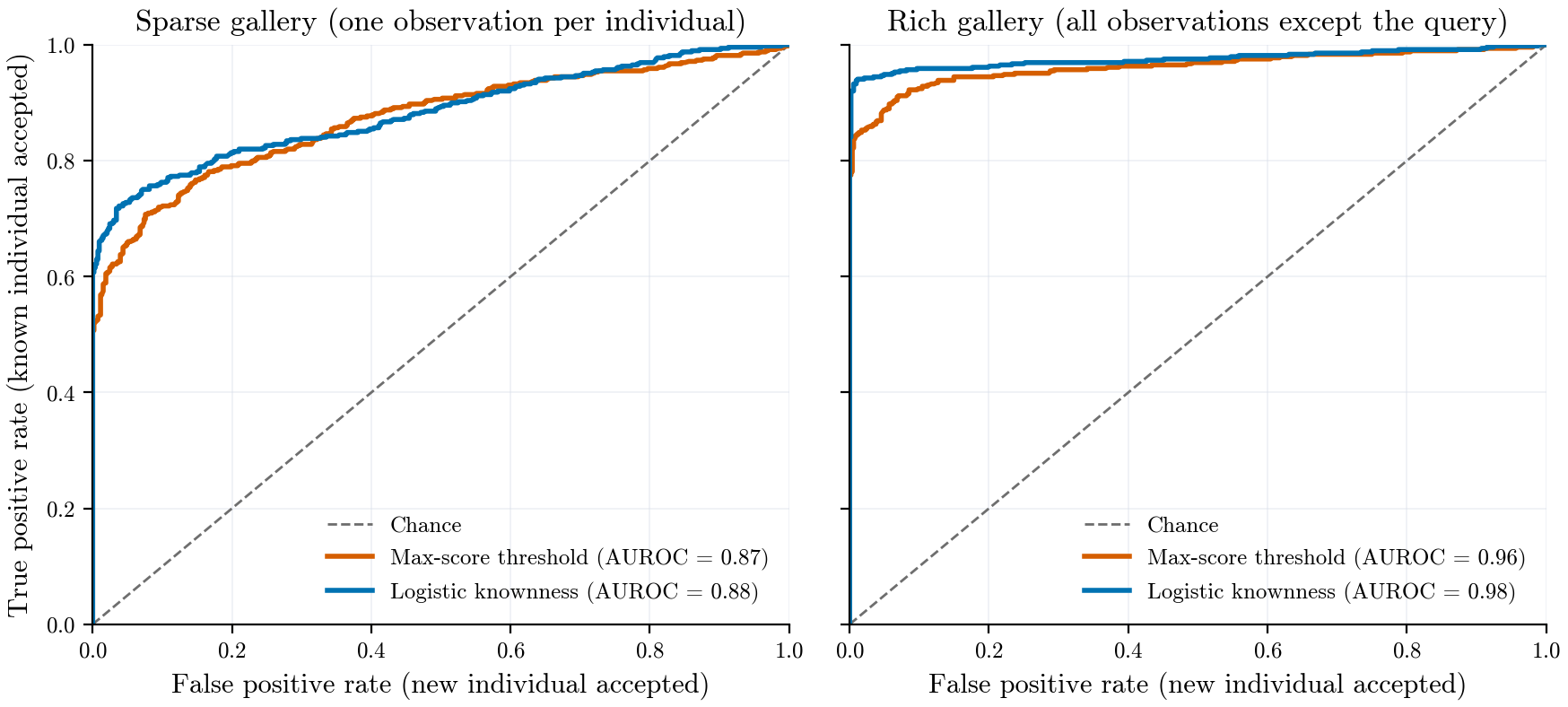}
\caption{True positive rate as a function of the false positive rate for determining whether the query corresponds to a new individual or an individual in the gallery.
The positive class is a known individual.
Comparison of maximum-score thresholding (orange curve, threshold applied to the pooled inlier count of step 5) and the logistic knownness detection model of step 6 (blue curve).}
\label{fig:open-set-detection}
\end{figure}

In both settings, the logistic knownness detection model distinguished new individuals slightly better than max-score thresholding (Figure~\ref{fig:open-set-detection}). Its AUROC was 0.88 against 0.87 with a sparse gallery, and 0.98 against 0.96 with a rich gallery.
Replacing the tuned threshold on $p_\text{known}$ by the MAP decision rule changed NA by 0.01 at most. 
 
Known individuals were harder to re-identify when the last gallery observation was older (Figure \ref{fig:baks_vs_time}).
With a sparse gallery, the full pipeline reached a BAKS of 0.92 [0.84, 0.98] when the gallery observation was at most a week away from the query.
When it was more than a year away, BAKS fell to 0.33 [0.19, 0.47]. 
With a rich gallery, most queries had an observation of the same individual less than three months earlier, and the BAKS of the full system stayed between 0.77 and 0.92 across the time bins.

\begin{table}[htbp]
\centering
\setlength{\tabcolsep}{4pt}
\resizebox{\textwidth}{!}{%
\begin{tabular}{lrrrrr}
\toprule
Method & BAKS $\uparrow$ & BAUS $\uparrow$ & BRA $\uparrow$ & NA $\uparrow$ & AUOSCR $\uparrow$ \\
\midrule
1. MiewID zero-shot & 0.23 [0.17, 0.29] (-) & 0.59 [0.52, 0.65] (-) & 0.45 [0.41, 0.50] (-) & 0.36 [0.31, 0.42] (-) & 0.23 [0.18, 0.29] (-) \\
2. + MiewID fine-tuning & 0.58 [0.50, 0.66] (+0.35) & 0.69 [0.63, 0.75] (+0.10) & 0.65 [0.60, 0.69] (+0.19) & 0.63 [0.58, 0.68] (+0.26) & 0.61 [0.53, 0.67] (+0.37) \\
3. + relevance weighting & 0.64 [0.56, 0.72] (+0.07) & 0.62 [0.55, 0.69] (-0.07) & 0.63 [0.58, 0.68] (-0.02) & 0.63 [0.59, 0.68] (+0.00) & 0.63 [0.56, 0.70] (+0.03) \\
4. + ALIKED/LightGlue rerank & 0.57 [0.49, 0.64] (-0.08) & 0.88 [0.83, 0.92] (+0.26) & 0.76 [0.72, 0.80] (+0.13) & 0.71 [0.65, 0.75] (+0.07) & 0.74 [0.68, 0.79] (+0.11) \\
5. + RANSAC verification & 0.73 [0.66, 0.79] (+0.16) & 0.85 [0.80, 0.89] (-0.03) & 0.80 [0.77, 0.83] (+0.04) & 0.78 [0.74, 0.82] (+0.08) & 0.80 [0.74, 0.84] (+0.06) \\
6. + logistic knownness & \textbf{0.75 [0.68, 0.81] (+0.02)} & 0.91 [0.88, 0.93] (+0.06) & 0.85 [0.82, 0.88] (+0.04) & \textbf{0.82 [0.78, 0.86] (+0.04)} & \textbf{0.82 [0.76, 0.87] (+0.02)} \\
7. + MAP open-set decision & 0.68 [0.61, 0.75] (-0.06) & \textbf{0.97 [0.96, 0.98] (+0.07)} & \textbf{0.87 [0.84, 0.90] (+0.02)} & \textbf{0.82 [0.77, 0.86] (-0.01)} & \textemdash{} \\
\bottomrule
\end{tabular}%
}
\caption{Seahorse re-ID with a sparse gallery. Cells are means over five gallery draws within the test split, with 95\% bootstrap intervals over test individuals in brackets and the change from the row above in parentheses.}
\label{tab:reid_progression_one_observation_per_individual}
\end{table}

\begin{table}[htbp]
\centering
\setlength{\tabcolsep}{4pt}
\resizebox{\textwidth}{!}{%
\begin{tabular}{lrrrrr}
\toprule
Method & BAKS $\uparrow$ & BAUS $\uparrow$ & BRA $\uparrow$ & NA $\uparrow$ & AUOSCR $\uparrow$ \\
\midrule
1. MiewID zero-shot & 0.38 [0.28, 0.48] (-) & 0.56 [0.48, 0.63] (-) & 0.49 [0.43, 0.55] (-) & 0.46 [0.39, 0.51] (-) & 0.36 [0.27, 0.44] (-) \\
2. + MiewID fine-tuning & 0.74 [0.65, 0.82] (+0.36) & 0.65 [0.57, 0.73] (+0.10) & 0.69 [0.63, 0.74] (+0.20) & 0.69 [0.64, 0.75] (+0.24) & 0.73 [0.66, 0.82] (+0.38) \\
3. + relevance weighting & 0.74 [0.65, 0.82] (-0.00) & 0.87 [0.82, 0.92] (+0.22) & 0.82 [0.78, 0.86] (+0.13) & 0.80 [0.75, 0.85] (+0.11) & 0.78 [0.70, 0.86] (+0.04) \\
4. + ALIKED/LightGlue rerank & 0.65 [0.56, 0.75] (-0.08) & 0.99 [0.98, 1.00] (+0.12) & 0.87 [0.83, 0.90] (+0.05) & 0.81 [0.74, 0.86] (+0.01) & 0.83 [0.75, 0.90] (+0.05) \\
5. + RANSAC verification & 0.79 [0.69, 0.87] (+0.13) & 0.94 [0.90, 0.97] (-0.05) & 0.88 [0.84, 0.92] (+0.02) & 0.86 [0.80, 0.91] (+0.05) & 0.86 [0.78, 0.93] (+0.03) \\
6. + logistic knownness & \textbf{0.85 [0.77, 0.93] (+0.06)} & 0.96 [0.93, 0.99] (+0.02) & 0.92 [0.89, 0.95] (+0.04) & 0.90 [0.86, 0.95] (+0.05) & \textbf{0.88 [0.80, 0.95] (+0.02)} \\
7. + MAP open-set decision & 0.82 [0.74, 0.90] (-0.03) & \textbf{1.00 [0.99, 1.00] (+0.04)} & \textbf{0.93 [0.90, 0.96] (+0.01)} & \textbf{0.91 [0.86, 0.95] (+0.00)} & \textemdash{} \\
\bottomrule
\end{tabular}%
}
\caption{Seahorse re-ID with a rich gallery. Cells are computed on the test split, with 95\% bootstrap intervals over test individuals in brackets and the change from the row above in parentheses. }
\label{tab:reid_progression_all_except_query}
\end{table}

\section{Discussion}


\subsection{Seahorse re-identification}
In this paper, we demonstrated the feasibility of high-performance automated seahorse taxonomic classification, sex determination and open-set individual re-identification from underwater photograph sequences in the wild. These are especially challenging tasks given that, in these conditions, image quality varies widely, and that European seahorses are cryptic animals with subtle patterns that rapidly change over time. To mimic either the continuation of the re-identification programme from which the dataset originated or, by contrast, the deployment of the system in a new population, we evaluated the re-ID models in rich- and sparse-gallery settings. Our results show that method performance can vary greatly depending on gallery composition. 

A preliminary study had reported promising performance in the automatic closed-set re-identification of \textit{H. guttulatus} from images \parencite{aloui_individual_2022}, with a top-1 closed-set accuracy of up to 76\% on manually curated data.
By contrast, our approach works in an open-set setting and reaches a BAKS of 82\% on uncurated data with a rich gallery. 


\subsection{Set-to-set re-identification}
We demonstrated that leveraging multiple photographs from underwater observations led to improved performance across these tasks. This is consistent with findings from the set-to-set face recognition literature \parencite{liu_quality_2017}.

Consistent with previous reports \parencite{yesharim_nearperfect_2026}, we found that fine-tuning of a re-identification foundation model was the most important step for performance improvement. Moreover, applying ALIKED/LightGlue reranking on a set of MiewID-shortlisted gallery identities clearly improved performance, with both gallery policies. Geometric verification with RANSAC further improved local feature matching. A separate "knownness detection" module gave a small gain over standard thresholding of the best match score with both gallery policies, but when combined with the MAP decision rule, it removed the need to tune a threshold.   

Two published methods that learn image weights without labels, QAN and PFE, did not improve the full system. The fine-tuned MiewID already separates the individuals from the training split quite well, and a linear probe on DINOv3 is able to classify images depending on their relevance to re-identification, which may have left them little margin for improvement.

\subsection{Limitations \& future work}

\paragraph{Re-identification of \textit{H. hippocampus}} The very small number of \textit{H. hippocampus} individuals in our dataset prevented us from evaluating the re-identification performance on this species specifically.
In future work, we aim to gather more \textit{H. hippocampus} observation data to provide meaningful performance estimates for each species.

\paragraph{Re-identification of juveniles and young adults}
\textcite{correia_crown_2021} reported that re-identification is more prone to errors for young adults, suggesting that the system's performance may not be consistent across all age categories in the seahorse population.
Our dataset contained a small fraction of re-identified juveniles, making it impractical to precisely estimate re-ID performance across life stages.

\paragraph{Transfer to other locations}
Significant morphological differences exist between populations of the same seahorse species \parencite{woodall_synthesis_2017, oruc_morphological_2024}.
Therefore, it is not clear how well the performance of our trained model will transfer to sites other than the Thau lagoon.
Site-specific fine-tuning may be needed to reach performance similar to that reported in this paper.

\paragraph{Transfer to citizen science programmes}
Our dataset is composed of photographs taken by marine biologists who are expert divers and marine photographers.
They worked in favourable conditions: at night, using flashes, and in some cases while handling the seahorse.
These conditions may not be as favourable in citizen science projects where recreational divers are involved, and it is not clear at this stage how performance would transfer.

\paragraph{Leveraging photograph metadata}

Given that seahorses are mostly sedentary and display strong site fidelity \parencite{caldwell_sedentary_2012, boehm_marine_2013, siqueira_estimating_2017}, photo location and date could serve as additional inputs to the re-identification model and may improve performance.

\ifarxiv
\section*{CRediT Authorship contribution statement}
Tristan Fauvel: Conceptualization, Methodology, Software, Validation, Formal analysis, Visualization, Project administration, Funding acquisition, Writing – original draft, Writing – review \& editing. Sylvie Louisy: Investigation, Resources, Data curation, Writing – review \& editing. Patrick Louisy: Investigation, Resources, Writing – review \& editing. Thomas Fauvel: Conceptualization, Project administration, Funding acquisition, Writing – review \& editing.
\fi

\section*{Declaration of generative AI and AI-assisted technologies in the manuscript preparation process}
  
During the preparation of this work, the authors used Claude Code (Opus 5.5) for code writing and ideation assistance, and GPT 5.6-Sol for proofreading. After using these tools, the authors reviewed and edited the content as needed and take full responsibility for the content of the published article.

\section*{Conflict of interest statement}
The authors declare no conflict of interest.

\section*{Data and code availability} 
\ifarxiv
The code used to produce our results is available at \url{https://github.com/TristanFauvel/seahorse-reidentification-notebooks}.
\else
The code used to produce our results is available at [link removed for anonymous review].
\fi  
The photographs are available upon request. 

\printbibliography

\clearpage
\appendix
\setcounter{table}{0}
\setcounter{figure}{0}
\setcounter{equation}{0}
\renewcommand{\thetable}{S\arabic{table}}
\renewcommand{\thefigure}{S\arabic{figure}}
\renewcommand{\theequation}{S\arabic{equation}}
\renewcommand{\theHtable}{S.\arabic{table}}
\renewcommand{\theHfigure}{S.\arabic{figure}}
\renewcommand{\theHequation}{S.\arabic{equation}}
\renewcommand{\thesection}{S\arabic{section}}
\renewcommand{\theHsection}{S.\arabic{section}}
\makeatletter
\renewcommand{\@seccntformat}[1]{\csname prefix@#1\endcsname\csname the#1\endcsname\quad}
\newcommand{\prefix@section}{Appendix~}
\begingroup
\title{Supporting Information: Automated open-set multi-view seahorse re-identification from underwater photograph sequences}
\author{}
\date{}
\thispagestyle{empty}
\@maketitle
\endgroup
\makeatother
\section{Supplementary methods}
\subsection{Late fusion rules for species and sex classification}
\label{sec:late_fusion_rules}

If we assume that the views are conditionally independent given the label, the likelihood is given by $p(X_{1:m}|Y) = \prod_{i=1}^m p(X_i|Y)$, where $X_{1:m}$ is an observation made of $m$ images and $Y$ is the observation label (either "$\text{new}$" or an individual in the gallery). This assumption means that, conditional on the class label, observing one view gives no additional information about any other view.

By denoting the labelled dataset $\mathcal{D}$, the posterior predictive class probability is given by:

\begin{align}
     p(Y=1|X_{1:m}, \mathcal{D}) &\propto P(Y=1|\mathcal{D})\prod_{i=1}^m p(X_i|Y=1, \mathcal{D}) \\
    &\propto p(Y=1|\mathcal{D})\prod_{i=1}^m \frac{P(Y=1|X_i, \mathcal{D})}{P(Y=1|\mathcal{D})}
\end{align} 
Let $L$ be the posterior log odds (or logit) of the observation, $L = \log \frac{P(Y=1|X_{1:m}, \mathcal{D})}{P(Y=0|X_{1:m}, \mathcal{D})}$.
We get: 
\begin{equation}
    L = \operatorname{logit}(\pi) + \sum_{i=1}^m \left [\operatorname{logit}(p_i) - \operatorname{logit}(\pi)\right]
\end{equation}
with the positive class marginal posterior probability $\pi = P(Y=1|\mathcal{D})$ and the posterior predictive probability for the positive class $p_i = P(Y=1|X_i, \mathcal{D})$.
The image score is the logit of the image-level classifier, so $\ell_i = f(X_i) = \operatorname{logit}(p_i)$.

We fitted the image-level classifier with weights that give the two classes the same total weight, so that $\pi = \frac{1}{2}$ and $\operatorname{logit}(\pi) = 0$.
Under the conditional independence assumption, we then get:
\begin{equation}
P(Y=1|\ell_{1:m}) = \sigma(L) = \sigma \left(\sum_{i=1}^m \ell_i \right)
\end{equation}
where $\sigma$ is the sigmoid function.

The assumption of conditional independence of the views is, of course, optimistic, as after conditioning on the class label, images from the same observation may remain dependent through shared observation conditions such as turbidity, lighting, or background.  

At the opposite extreme, one may consider the limiting case of complete redundancy, in which any single view contains all the class-relevant information available in the entire set of views.
This assumption can be expressed as
$P(Y=1 \mid X_{1:m},\mathcal{D}) = 
P(Y=1 \mid X_i,\mathcal{D}).$

Requiring this relation to hold for every $i$ is a strong assumption: it implies that each individual view contains the same class-relevant information as the complete set.  
Under this assumption, we obtain the average-logit fusion rule: 
\begin{equation}
P(Y=1|\ell_{1:m}, \mathcal{D}) = \sigma \left(\frac{1}{m}\sum_{i=1}^m \ell_i \right)
\end{equation}
Note that, in particular, adding duplicate images does not increase the magnitude of the fused logit and therefore does not produce a more confident posterior. In practice, one may want to
correct for redundancy across images to prevent an overconfident posterior, by simply using: 
 
 \begin{equation}
P(Y=1|\ell_{1:m}, \mathcal{D}) = \sigma \left(\alpha \sum_{i=1}^m \ell_i \right)
\end{equation}
where $0<\alpha<1$.
Note that this scaling factor does not affect the decision boundary, hence the predicted classes.
The average-logit fusion rule corresponds to a conservative choice of $\alpha = m^{-1}$.

\subsection{Re-identification methods}
\subsubsection{MiewID fine-tuning}
\label{app:miewid_finetuning}

We replaced MiewID's original classification head with a sub-centre ArcFace head \parencite{deng_arcface_2019} over the seahorse identities, with three sub-centres, a scale of 64, and a margin of 0.5 rad. We trained all backbone layers except the BatchNorm modules, which we kept frozen throughout training.

We used the AdamW optimiser with differential learning rates (backbone: 2e-5, ArcFace head: 1e-3, weight decay 1e-4), and a cosine annealing learning rate scheduler over 100 epochs down to a minimum learning rate of 1e-6. 
Moreover, we applied data augmentation, including random resized crops (scale 0.75–1.0), colour jittering ($p=0.8$), random greyscale conversion ($p=0.2$), and Gaussian blur ($p=0.5$).
We used a batch size of 24, comprising four images from each of six identities. 

Every five epochs, we computed mAP@R on the known queries of the validation split, and we kept the checkpoint with the highest value. We split the validation individuals into five groups, stratified by observation count. In round i, the queries were the observations of group i + 1 individuals with at least two observations. The gallery held all validation observations except those of group i and the query's own observation. We averaged mAP@R over the five rounds.

\subsubsection{Derivation of the open-set decision rule}
\label{app:open-set_decision_rule}
We start from the Bayes-optimal decision rule (equation~\mainref{eq:bayes_optimal_rule}), with the notation of Section~\mainref{sec:bayes_rule}.

We have: 
\begin{equation}
 \begin{aligned}
 \underset{q \in \mathcal{G} \cup \{\text{new}\} }{\max}P(Y=q|X_{1:m}, \mathcal{D}) &= \max \left[ \underset{q \in \mathcal{G} }{\max \,}P(Y=q|X_{1:m}, \mathcal{D}), P(Y = \text{new}|X_{1:m}, \mathcal{D}) \right]
\end{aligned}
\end{equation}

Let $K = \mathds{1}\{q\in \mathcal{G}\}$ denote the binary variable that indicates whether the individual in the query belongs to the gallery ($K=1$) or not ($K=0$).
The probability that an observation corresponds to individual $q$ in the gallery is given by: 

\begin{equation}
 \begin{aligned}
P(Y=q|X_{1:m}, \mathcal{D})
 &= P(K=1|X_{1:m}, \mathcal{D})P(Y=q|X_{1:m}, K=1, \mathcal{D}) \\
 &= p_{\text{known}}s_q,
\end{aligned}
\end{equation}
where $s_{q}$ is the closed-set re-ID probability, and $p_\text{known}$ is the probability that the query individual belongs to the gallery.

Moreover, we have: 
\begin{equation}
 \begin{aligned}
    P(Y = \text{new}|X_{1:m}, \mathcal{D}) &= P(K = 0|X_{1:m}, \mathcal{D}) \\
    &= 1 - p_{\text{known}}
 \end{aligned}
 \end{equation}
So that: 

\begin{equation}
 \begin{aligned}
 \underset{q \in \mathcal{G} \cup \{\text{new}\} }{\max}P(Y=q|X_{1:m}, \mathcal{D}) &= \max \left[ 1 - p_{\text{known}}, p_{\text{known}}\max_{q\in\mathcal{G}}s_q \right]
\end{aligned}
\end{equation}

The prediction is therefore:
\begin{equation}
\begin{aligned}
\hat{q} &= \underset{q \in \mathcal{G} \cup \{\text{new}\} }{\operatorname{argmax}}P(Y=q|X_{1:m}, \mathcal{D}) \\ 
&= \begin{cases}
 \underset{q \in \mathcal{G} }{\operatorname{argmax}} s_q & \text{if } \frac{1}{\max_{q\in\mathcal{G}}s_q + 1}<p_\text{known},\\
\text{new}, & \text{otherwise}.
\end{cases} 
\end{aligned}
\label{eq:decision_rule}
\end{equation}
 An observation is therefore more likely to be considered as from a new individual as $\max_{q\in\mathcal{G}}s_q$ decreases.
 There is no fixed threshold $\tau$ such that the decision rule systematically assigns an individual in the gallery if $p_\text{known} > \tau$ or if $\max_{q\in\mathcal{G}}s_q > \tau$. 
Note that $\frac{1}{2}\leq \frac{1}{\max_{q\in\mathcal{G}}s_q + 1}<1$ and the decision rule can assign the query observation to an individual in the gallery only if $p_\text{known} > \frac{1}{2}$.

Therefore, for each observation, we need to estimate two quantities: the probability that the observation corresponds to a known individual, $p_\text{known}=P(K=1|X_{1:m}, \mathcal{D})$, and the closed-set re-identification probability $s_q = P(Y=q|X_{1:m}, K=1, \mathcal{D})$. 
The closed-set re-identification decision is denoted $\hat{z} = \underset{q \in \mathcal{G} }{\operatorname{argmax}} P(Y=q|X_{1:m}, K=1, \mathcal{D}) $.
The knownness decision is given by:

\begin{equation}
\label{eq:knownness_decision}
\begin{aligned}
\hat{\gamma} &= \mathds{1}\left\{\frac{1}{\max_{q\in\mathcal{G}}s_q + 1}<p_\text{known}\right\} \\
&= \mathds{1}\left\{\frac{1}{s_{\hat{z}} + 1}<p_\text{known}\right\}
\end{aligned}
\end{equation}

So that: 
  
\begin{equation}
\begin{aligned}
\hat{q} &=     
\begin{cases}
\hat{z}, & \text{if } \hat{\gamma} = 1,\\
\text{new}, & \text{if } \hat{\gamma} = 0.
\end{cases}
\end{aligned}
\end{equation}

\subsubsection{Re-identification metrics}
\label{app:reid_metrics}

The Balanced Re-identification Accuracy (BRA) is defined in Equation~\mainref{eq:bra} of Section~\mainref{sec:reid_metrics}.
In this equation, $\rho$ is the share of known-query individuals among the query individuals of both roles.
An individual that is a query in both roles counts once in each term.

BAUS is the Balanced Accuracy on Unknown Samples \parencite{adam_wildlifereid10k_2025}, defined as:
\begin{equation} 
\mathrm{BAUS} = \frac{1}{\left|\bar{\mathcal{G}}\cap \mathcal{Q}\right|} \sum_{q \in \bar{\mathcal{G}}\cap \mathcal{Q}} \frac{1}{\left|\mathcal{O}_{q}\right|} \sum_{i \in \mathcal{O}_{q}} \mathds{1}\left(\hat{q}_i=\text{new}\right)
\end{equation}

BAKS is the Balanced Accuracy on Known Samples \parencite{adam_wildlifereid10k_2025}:
\begin{equation} 
\mathrm{BAKS} = \frac{1}{\left|\mathcal{G} \cap \mathcal{Q}\right|} \sum_{q \in \mathcal{G}\cap \mathcal{Q}} \frac{1}{\left|\mathcal{O}_{q}\right|} \sum_{i \in \mathcal{O}_{q}} \mathds{1}\left(\hat{q}_i=q\right)
\end{equation}
BAKS answers the question: "for a known animal, did the full open-set system both accept it as known and choose the right top-1 identity?".

A system that classifies every query as a new individual would get a BAUS of $1$, and a BRA of $1-\rho$.
Therefore, it is useful to consider the normalised accuracy, which is the geometric mean of BAKS and BAUS \parencite{adam_wildlifereid10k_2025}: 

\begin{equation}
    \text{Normalised Accuracy} = \sqrt{\mathrm{BAKS} \cdot \mathrm{BAUS}} 
\end{equation}

Let $\mathrm{BAKS}(\tau)$ and $\mathrm{BAUS}(\tau)$ be the two metrics when every query with a decision score of at least $\tau$ is accepted as known.
The pairs $(\mathrm{BAKS}(\tau), \mathrm{BAUS}(\tau))$ trace a balanced open-set classification rate (OSCR) curve \parencite{dhamija_reducing_2018}.
The area under this curve is:
\begin{equation}
\mathrm{OSCR} = \int \mathrm{BAUS} \, \mathrm{d}\,\mathrm{BAKS} = \frac{1}{\left|\mathcal{G} \cap \mathcal{Q}\right|} \sum_{q \in \mathcal{G}\cap \mathcal{Q}} \frac{1}{\left|\mathcal{O}_{q}\right|} \sum_{i \in \mathcal{O}_{q}} \mathds{1}\left(\hat{z}_i=q\right) \mathrm{BAUS}(c_i),
\label{eq:OSCR}
\end{equation}
where $c_i$ is the decision score of query observation $i$.
In $\mathrm{BAUS}(c_i)$, an unknown query with the same score as $c_i$ counts as half rejected. AUOSCR is at most the closed-set balanced accuracy (BAKS).

\subsubsection{Known/unknown probability model}
\label{app:known_unknown_model}

The knownness model (Section~\mainref{sec:modelling_approaches}) takes the following features as input:

\begin{itemize}
    \item Local features matching confidence gap between the best and second-best identities, measured as the gap between the best and second-best identity's pooled ALIKED/LightGlue inlier count.
    A large gap means one identity is clearly matched; a small gap means the query resembles no gallery individual decisively (typical of a genuinely new animal). 
    \item Fraction of the photographs in an observation whose own image-level MiewID top-1 identity agrees with the observation's best matching identity.
    Concretely, query image/gallery identity similarity score was computed by max-pooling, and the top-1 match for each query image $i$, $\underset{g}{\argmax}\underset{j}{\max \,} S_C(X_i, X^{(g)}_j)$, was returned.
    The fraction of image-level top-1 matches corresponding to the observation-level top-1 match measures the within-observation consistency. 
    \item Mean similarity to the winning identity, computed as the average image-level cosine similarity derived from the fine-tuned MiewID from the query observation's photographs to the best matching identity's photographs determined by local feature matching $\hat{z}$, $\frac{1}{m} \sum_{i=1}^m \underset{j}{\max \,} S_C(X_i, X^{(\hat{z})}_j) $.
    This gives a measure of how strongly, on average, the MiewID embeddings support the best match.
    \item Standard deviation of the similarity to the winning identity, computed as the standard deviation of the image-level cosine similarity derived from the fine-tuned MiewID from the query observation's photographs to the best matching identity's photographs determined by local feature matching, $\operatorname{std}_{i=1:m} \left[ \underset{j}{\max \,} S_C(X_i, X^{(\hat{z})}_j )\right]$. 
    \item Binary variable corresponding to the agreement between the ALIKED/LightGlue rerank and the fine-tuned MiewID top match.
    Intuitively, when global embedding and local descriptors agree, the "known" decision is more trustworthy.

\end{itemize}

\subsubsection{Open-set MAP decision rule}
\label{app:map_open_set_rule}

First, we Platt-scaled the local-feature matching scores into
matching probabilities.
Let $c$ be the pooled ALIKED/LightGlue score of a (query observation, shortlisted identity) pair, that is, the mean of its five highest image-pair inlier counts.
We fitted a one-dimensional logistic regression on $\log(1+c)$, over all shortlisted pairs from the fitting fraction of the validation split.
Each pair was labelled by whether the query shows that identity.
We write $P_\text{match}(X_{1:m},k)$ for the resulting probability that the query observation $X_{1:m}$ shows identity $k$.
Non-shortlisted identities received $P_\text{match}=0$.
We hypothesised that this simplifying approximation would have little impact, given the relatively large shortlist.
To obtain the closed-set re-ID probability distribution from these matching probabilities, we then normalised over the gallery identities:
\begin{equation}
  P(Y=k \mid X_{1:m}, K{=}1) \;=\;
  \frac{P_\text{match}(X_{1:m},k)}{\sum_{g \in \mathcal{G}} P_\text{match}(X_{1:m},g)} .
\end{equation}

$P_\text{match}$ is calibrated, but $s_{\hat{z}} = \max_{k \in \mathcal{G}}
P(Y=k \mid X_{1:m}, K{=}1)$, the probability of the mode $\hat{z}$, is not.
We therefore applied temperature scaling.
We raised each probability in the distribution to a power $T$ and normalised again over $\mathcal{G}$, so the result was still a distribution.
We fitted $T$ by minimising the log-loss of the scaled $s_{\hat{z}}$ against top-1 match correctness.
The fit used the known queries of the scoring fraction of the validation split.

The second probability in the MAP open-set decision rule, $p_\text{known}$, also needs recalibration as it is the result of a logistic regression fitted with balanced known/unknown weights and identity weights. We therefore Platt-recalibrated it against gallery membership on the scoring fraction of the validation split.

\subsection{Benchmark methods for image weighting}
\label{sec:supp_methods}

We compared relevance weighting (step 3) with two published methods that also weight the photographs of a set: QAN \parencite{liu_quality_2017} and PFE \parencite{shi_probabilistic_2019}. Both methods replace step 3 only, and both use the fine-tuned MiewID of step 2. Steps 4 to 7 are unchanged and use the step-3 scores.

\paragraph{Quality Aware Network}
QAN gives each image a quality score between 0 and 1, and the feature of a set (i.e. an observation) is the quality-weighted mean of its image features \parencite{liu_quality_2017}.
We kept the fine-tuned MiewID frozen, so that only the aggregation method differs between the compared methods. We added a quality unit made of one linear layer followed by a sigmoid, which takes the MiewID embedding of each photograph as input. We used a batch-hard triplet loss on the set features, with a margin of 0.3 in cosine distance. For each batch, we drew 16 individuals with at least two observations, up to 4 observations of each individual, and a random subset of 1 to 8 photographs of each observation. The weights of the quality unit were initialised at zero, so every photograph started at quality 0.5. 
We used the Adam optimiser (learning rate $10^{-3}$, weight decay $10^{-4}$) for 2,000 steps.
We also evaluated QAN pooling with the same quality for every photograph.
With this variant, we measured the effect of  pooling features instead of scores (equation~\mainref{eq:image_aggregation_baseline}).

\paragraph{Probabilistic Face Embeddings}
PFE treats each embedding $\mu$ as the mean of a Gaussian, and an uncertainty module predicts a variance $\sigma^2_l$ for each dimension $l$ \parencite{shi_probabilistic_2019}.
Two Gaussians are compared by the mutual likelihood score (MLS), the log-likelihood that both come from the same latent embedding.
Up to a positive factor and a constant, it is:
\begin{equation}
\mathrm{MLS}(a, b) = -\sum_{l=1}^{D} \left[ \frac{(\mu_{a,l} - \mu_{b,l})^2}{\sigma^2_{a,l} + \sigma^2_{b,l}} + \log\left(\sigma^2_{a,l} + \sigma^2_{b,l}\right) \right],
\label{eq:mls}
\end{equation}
with $D = 2152$ for MiewID.
The loss is the mean negative MLS over genuine pairs.

For each dimension, we weighted the photographs by the inverse of their variance to get the fused mean, and then normalised this mean to unit length:
\begin{equation}
\hat{\mu}_l \propto \frac{\sum_{i=1}^m \mu_{i,l}/\sigma^2_{i,l}}{\sum_{i=1}^m 1/\sigma^2_{i,l}}, \qquad \hat{\sigma}^2_l = \min_{i} \sigma^2_{i,l}.
\end{equation}

We followed the authors' code release for the architecture, the training schedule and the fusion.
The input of the module is the fine-tuned MiewID embedding before its L2 normalisation.
Each batch held 4 photographs from each of 64 individuals.
We used stochastic gradient descent with momentum 0.9 and weight decay $5\times10^{-4}$ for 12,000 steps, and divided the learning rate by 10 at step 8,000.
The learning rate used in the PFE authors' code is $10^{-3}$ for 512-dimensional embeddings. Since the loss is a sum over dimensions, we multiplied the learning rate by $512/D$.
Without this scaling, training diverged. 
 

\section{Supplementary results}

\subsection{Species classification}
\label{sec:supp_species}

DINOv3 linear probing led to better performance than BioCLIP 2.5 in terms of balanced accuracy, macro F1 and AUPRC (Table \ref{tab:species_backbone}), both at the image level and at the observation level with average-logit fusion (Note that \textit{H. hippocampus} is the positive class). With DINOv3, all three fusion rules raised balanced accuracy from 0.93 to 1.00, whereas the AUPRC rose from 0.86 to 1.00 (Table \ref{tab:species_fusion}). All three fusion rules lowered the Brier score and the Expected Calibration Error (ECE). Average-logit fusion and majority vote lowered the Maximum Calibration Error (MCE), and logit-sum fusion raised it to 0.74.

These results rest on only 13 \textit{H. hippocampus} individuals and 48 observations, which is too few to reliably estimate species classification performance. With average-logit fusion, the classifier assigned all 48  \textit{H. hippocampus} observations to the correct species. It assigned 3
of the 2,437 \textit{H. guttulatus} observations, from 3 of the 460 individuals, to \textit{H. hippocampus}. With 13 \textit{H. hippocampus} individuals, a bootstrap over individuals is not reliable. With individuals as units, the exact 95\% Clopper--Pearson interval of the \textit{H. hippocampus} recall is [0.75, 1.00].

\begin{table}[htbp]
\centering
\begin{tabular}{lrrr}
\toprule
Method & Balanced acc. $\uparrow$ & Macro F1 $\uparrow$ & AUPRC $\uparrow$ \\
\midrule
DINOv3 ViT-L/16 CLS (image) & 0.93 & 0.84 & 0.86 \\
DINOv3 ViT-L/16 CLS (observation) & \textbf{1.00} & \textbf{1.00} & \textbf{1.00} \\
BioCLIP 2.5 (image) & 0.84 & 0.66 & 0.50 \\
BioCLIP 2.5 (observation) & 0.93 & 0.89 & 0.89 \\
\bottomrule
\end{tabular}
\caption{Backbone comparison at the image and observation levels (observation = average-logit fusion) for species classification.
All metrics are identity-balanced, over the 13 \textit{H. hippocampus} and 460 \textit{H. guttulatus} individuals.
AUPRC: Area Under the Precision-Recall Curve, with \textit{H. hippocampus} as the positive class.}
\label{tab:species_backbone}
\end{table}

\begin{table}[htbp]
\centering
\begin{tabular}{lrrrrrr}
\toprule
Method & Balanced acc. $\uparrow$ & Macro F1 $\uparrow$ & AUPRC $\uparrow$ & Brier $\downarrow$ & ECE $\downarrow$ & MCE $\downarrow$ \\
\midrule
Image (LR) & 0.93 & 0.84 & 0.86 & 0.02 & 0.05 & 0.52 \\
Average logit & \textbf{1.00} & \textbf{1.00} & \textbf{1.00} & 3.40${}\times 10^{-3}$ & 0.03 & \textbf{0.41} \\
Logit sum & \textbf{1.00} & \textbf{1.00} & \textbf{1.00} & \textbf{3.61${}\times 10^{-4}$} & \textbf{4.66${}\times 10^{-4}$} & 0.74 \\
Majority vote & \textbf{1.00} & 0.99 & \textbf{1.00} & 3.55${}\times 10^{-3}$ & 0.02 & 0.47 \\
\bottomrule
\end{tabular}
\caption{Comparison of image fusion methods with DINOv3 CLS, with the per-image logistic regression as a baseline, in species classification. All metrics are identity-balanced. We computed the calibration errors with ten equal-width probability bins. AUPRC: Area Under the Precision-Recall Curve, with \textit{H. hippocampus} as the positive class ECE: Expected Calibration Error. MCE: Maximum Calibration Error.}
\label{tab:species_fusion}

\end{table}

\subsection{Sex classification}

Table \ref{tab:sex_fusion} gives the sex determination results described in Section \mainref{sec:classification_results}.

\begin{table}[htbp]
\centering
\setlength{\tabcolsep}{4pt}
\resizebox{\textwidth}{!}{%
\begin{tabular}{lrrrrrr}
\toprule
Method & Balanced acc. $\uparrow$ & Macro F1 $\uparrow$ & AUPRC $\uparrow$ & Brier $\downarrow$ & ECE $\downarrow$ & MCE $\downarrow$ \\
\midrule
Image (LR) & 0.80 [0.78, 0.82] & 0.80 [0.78, 0.82] & 0.90 [0.88, 0.92] & 0.14 [0.13, 0.15] & \textbf{0.04 [0.03, 0.06]} & \textbf{0.08 [0.06, 0.15]} \\
Average logit & \textbf{0.90 [0.87, 0.92]} & \textbf{0.89 [0.87, 0.92]} & \textbf{0.97 [0.96, 0.98]} & \textbf{0.08 [0.07, 0.09]} & 0.06 [0.05, 0.09] & 0.21 [0.18, 0.30] \\
Logit sum & \textbf{0.90 [0.87, 0.92]} & \textbf{0.89 [0.87, 0.92]} & \textbf{0.97 [0.95, 0.98]} & 0.09 [0.07, 0.12] & 0.09 [0.07, 0.12] & 0.64 [0.44, 0.76] \\
Majority vote & 0.88 [0.86, 0.91] & 0.88 [0.85, 0.91] & 0.95 [0.92, 0.97] & 0.09 [0.08, 0.11] & 0.05 [0.04, 0.08] & 0.19 [0.15, 0.26] \\
\bottomrule
\end{tabular}%
}
\caption{Comparison of image fusion methods with DINOv3 CLS, with the per-image logistic regression as a baseline, in sex classification.
All metrics are identity-balanced, with 95\% individual-level bootstrap intervals. AUPRC: Area Under the Precision-Recall Curve, with males as the positive class.
ECE: Expected Calibration Error.
MCE: Maximum Calibration Error.}
\label{tab:sex_fusion}
\end{table}

\clearpage
\subsection{Re-identification from single photographs vs from whole observations}
\label{ref:reid_single_vs_multi}

Table \ref{tab:reid_query_granularity_finetuned} compares the re-identification performance when using the fine-tuned MiewID model either at image-level or using our observation-level approach (step 2 of the pipeline).

\begin{table}[htbp]
\centering
\setlength{\tabcolsep}{4pt}
\resizebox{\textwidth}{!}{%
\begin{tabular}{llrrrrr}
\toprule
Gallery & Query unit & BAKS $\uparrow$ & BAUS $\uparrow$ & BRA $\uparrow$ & NA $\uparrow$ & AUOSCR $\uparrow$ \\
\midrule
Sparse & Single photograph & 0.33 [0.28, 0.38] (-) & 0.67 [0.63, 0.71] (-) & 0.54 [0.52, 0.58] (-) & 0.47 [0.43, 0.51] (-) & 0.34 [0.29, 0.38] (-) \\
Sparse & Observation & \textbf{0.58 [0.50, 0.66] (+0.25)} & \textbf{0.69 [0.63, 0.75] (+0.02)} & \textbf{0.65 [0.60, 0.69] (+0.10)} & \textbf{0.63 [0.58, 0.68] (+0.16)} & \textbf{0.61 [0.53, 0.67] (+0.27)} \\
Rich & Single photograph & 0.44 [0.39, 0.50] (-) & \textbf{0.76 [0.72, 0.79] (-)} & 0.64 [0.61, 0.67] (-) & 0.58 [0.54, 0.62] (-) & 0.47 [0.42, 0.53] (-) \\
Rich & Observation & \textbf{0.74 [0.65, 0.82] (+0.30)} & 0.65 [0.57, 0.73] (-0.10) & \textbf{0.69 [0.63, 0.74] (+0.04)} & \textbf{0.69 [0.64, 0.75] (+0.12)} & \textbf{0.73 [0.66, 0.82] (+0.26)} \\
\bottomrule
\end{tabular}%
}
\caption{Re-identification from single photographs and from whole observations (equation \mainref{eq:image_aggregation_baseline}), with the fine-tuned MiewID descriptor (step 2 of Tables \mainref{tab:reid_progression_one_observation_per_individual} and \mainref{tab:reid_progression_all_except_query}). The change in parentheses is the observation row minus the single-photograph row.}
\label{tab:reid_query_granularity_finetuned}
\end{table}

\subsection{Re-identification performance as a function of the time between observations}
\label{sec:supp_time}

\begin{figure}[htbp]
\centering
\includegraphics[width=0.75\linewidth]{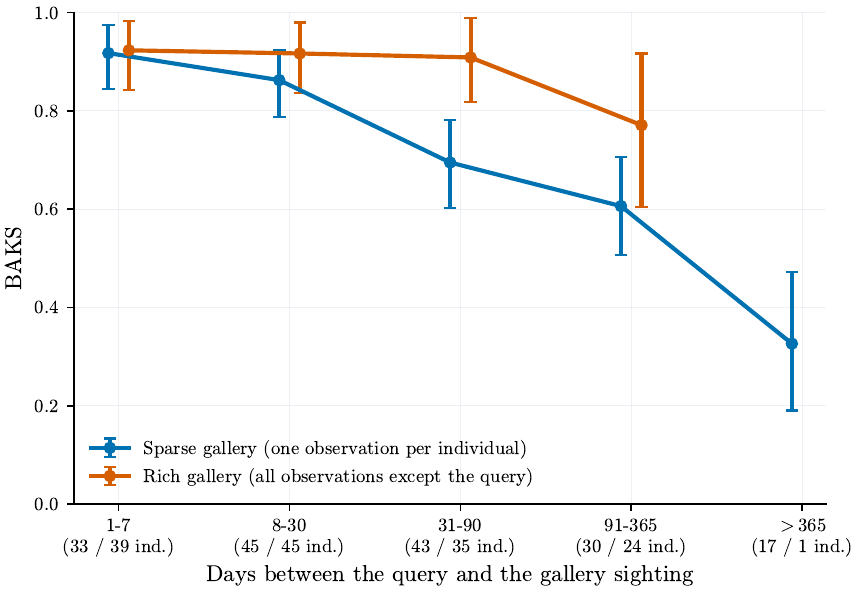}
\caption{BAKS of the full re-identification pipeline against the time between two observations of the same individual, on the test split. With a sparse gallery, the time is the gap between the query and the one gallery observation of the same individual, before or after the query. With a rich gallery, the time is the gap to the last earlier sighting, and queries with no earlier sighting are left out. Error bars are 95\% bootstrap intervals over individuals. A bin with fewer than three individuals is not plotted.}
\label{fig:baks_vs_time}
\end{figure}

\subsection{Benchmark results}
\label{sec:supp_results}


Pooling features rather than scores raised NA by 0.04 with a sparse gallery and by 0.06 with a rich gallery (Table~\ref{tab:supp_step3_benchmark}, QAN pooling, equal quality). The learned quality changed NA by 0.01 at most compared with equal quality. Used alone, QAN pooling beats relevance weighting with a sparse gallery, but not with a rich gallery. 

Used alone, PFE with the MLS gave the lowest NA of all step-3 methods with a rich gallery (Table~\ref{tab:supp_step3_benchmark}).
  
PFE fusion compared by cosine lowered BAUS under both gallery policies, compared with QAN pooling with equal quality: by 0.13 with a sparse gallery and by 0.05 with a rich gallery.
It changed BAKS by 0.07 at most.

With steps 4 to 7 added, the five choices of step 3 gave NA values within 0.01 of each other with a sparse gallery, and within 0.04 with a rich gallery, and all their intervals largely overlap (Table~\ref{tab:supp_late_steps}).
At step 7, PFE with the MLS gave the lowest BAUS and the lowest BAKS with a rich gallery. With a sparse gallery, PFE with cosine similarity gave the lowest BAUS (0.95), and all five choices gave a BAKS of 0.68. With a rich gallery, PFE with cosine similarity gave the highest BRA (0.94). One reason that may explain why the learned weights brought no gain in terms of BAKS is that these methods were trained on the training split, whose individuals the fine-tuned MiewID already separates well.

\begin{table}[htbp]
\centering
\setlength{\tabcolsep}{4pt}
\resizebox{\textwidth}{!}{%
\begin{tabular}{llrrrrr}
\toprule
Gallery & Method & BAKS $\uparrow$ & BAUS $\uparrow$ & BRA $\uparrow$ & NA $\uparrow$ & AUOSCR $\uparrow$ \\
\midrule
Sparse & Score pooling (step 2) & 0.58 [0.50, 0.66] & 0.69 [0.63, 0.75] & 0.65 [0.60, 0.69] & 0.63 [0.58, 0.68] & 0.61 [0.53, 0.67] \\
Sparse & QAN pooling, equal quality & 0.62 [0.53, 0.69] & 0.74 [0.68, 0.79] & \textbf{0.69 [0.65, 0.73]} & \textbf{0.67 [0.62, 0.72]} & 0.64 [0.56, 0.70] \\
Sparse & QAN pooling, learned quality & 0.63 [0.55, 0.71] & 0.73 [0.68, 0.78] & \textbf{0.69 [0.65, 0.73]} & \textbf{0.67 [0.62, 0.72]} & \textbf{0.66 [0.58, 0.72]} \\
Sparse & PFE fusion, cosine & 0.55 [0.46, 0.64] & 0.61 [0.54, 0.68] & 0.59 [0.54, 0.64] & 0.58 [0.53, 0.63] & 0.56 [0.49, 0.64] \\
Sparse & PFE fusion, MLS & 0.32 [0.24, 0.40] & \textbf{0.89 [0.85, 0.93]} & 0.68 [0.64, 0.72] & 0.53 [0.47, 0.59] & 0.47 [0.40, 0.54] \\
Sparse & Relevance-weighted score pooling & \textbf{0.64 [0.56, 0.72]} & 0.62 [0.55, 0.69] & 0.63 [0.58, 0.68] & 0.63 [0.59, 0.68] & 0.63 [0.56, 0.70] \\
\midrule
Rich & Score pooling (step 2) & 0.74 [0.65, 0.82] & 0.65 [0.57, 0.73] & 0.69 [0.63, 0.74] & 0.69 [0.64, 0.75] & 0.73 [0.66, 0.82] \\
Rich & QAN pooling, equal quality & 0.71 [0.61, 0.80] & 0.81 [0.74, 0.87] & 0.77 [0.72, 0.82] & 0.75 [0.70, 0.81] & 0.80 [0.73, 0.87] \\
Rich & QAN pooling, learned quality & \textbf{0.80 [0.71, 0.88]} & 0.72 [0.64, 0.79] & 0.75 [0.69, 0.80] & 0.76 [0.71, 0.81] & \textbf{0.81 [0.74, 0.89]} \\
Rich & PFE fusion, cosine & 0.70 [0.60, 0.79] & 0.76 [0.68, 0.82] & 0.74 [0.68, 0.79] & 0.73 [0.67, 0.78] & 0.76 [0.69, 0.83] \\
Rich & PFE fusion, MLS & 0.42 [0.31, 0.53] & \textbf{0.90 [0.85, 0.94]} & 0.72 [0.67, 0.77] & 0.62 [0.53, 0.68] & 0.56 [0.47, 0.66] \\
Rich & Relevance-weighted score pooling & 0.74 [0.65, 0.82] & 0.87 [0.82, 0.92] & \textbf{0.82 [0.78, 0.86]} & \textbf{0.80 [0.75, 0.85]} & 0.78 [0.70, 0.86] \\
\bottomrule
\end{tabular}%
}
\caption{Comparison of QAN and PFE against relevance weighting, under both gallery policies. For all rows, we used the fine-tuned MiewID embeddings and applied a threshold to the best identity score to determine if the observation corresponds to a new individual or not. Score pooling (step 2) averages the image scores (equation~\mainref{eq:image_aggregation_baseline}). QAN pooling averages the image features, with equal quality or with the learned QAN quality. In PFE fusion, we weighted each dimension of the mean embedding by the inverse of the variance from the uncertainty module. The fused observations are compared by cosine similarity or by the mutual likelihood score (MLS). Relevance-weighted score pooling is step 3 in our pipeline.Brackets give 95\% bootstrap intervals over test individuals.}
\label{tab:supp_step3_benchmark}
\end{table}

\begin{table}[htbp]
\centering
\setlength{\tabcolsep}{4pt}
\resizebox{\textwidth}{!}{%
\begin{tabular}{llrrrrr}
\toprule
Gallery & Step 3 & BAKS $\uparrow$ & BAUS $\uparrow$ & BRA $\uparrow$ & NA $\uparrow$ & AUOSCR (step 6) $\uparrow$ \\
\midrule
Sparse & Relevance weighting & \textbf{0.68 [0.61, 0.75]} & 0.97 [0.96, 0.98] & \textbf{0.87 [0.84, 0.90]} & \textbf{0.82 [0.77, 0.86]} & \textbf{0.82 [0.76, 0.87]} \\
Sparse & QAN, learned quality & \textbf{0.68 [0.60, 0.75]} & \textbf{0.98 [0.97, 0.99]} & \textbf{0.87 [0.84, 0.90]} & 0.81 [0.77, 0.86] & 0.80 [0.73, 0.86] \\
Sparse & QAN, equal quality & \textbf{0.68 [0.61, 0.75]} & 0.97 [0.96, 0.98] & \textbf{0.87 [0.83, 0.90]} & \textbf{0.82 [0.77, 0.86]} & 0.80 [0.73, 0.86] \\
Sparse & PFE, cosine & \textbf{0.68 [0.61, 0.75]} & 0.95 [0.93, 0.98] & 0.85 [0.82, 0.89] & 0.81 [0.76, 0.85] & 0.80 [0.73, 0.85] \\
Sparse & PFE, MLS & \textbf{0.68 [0.61, 0.75]} & 0.96 [0.94, 0.97] & 0.85 [0.82, 0.88] & 0.81 [0.76, 0.85] & 0.77 [0.70, 0.83] \\
\midrule
Rich & Relevance weighting & 0.82 [0.74, 0.90] & \textbf{1.00 [0.99, 1.00]} & 0.93 [0.90, 0.96] & \textbf{0.91 [0.86, 0.95]} & 0.88 [0.80, 0.95] \\
Rich & QAN, learned quality & 0.82 [0.73, 0.90] & \textbf{1.00 [0.99, 1.00]} & 0.93 [0.89, 0.96] & 0.90 [0.85, 0.95] & 0.89 [0.83, 0.95] \\
Rich & QAN, equal quality & 0.83 [0.75, 0.91] & 0.98 [0.96, 1.00] & 0.93 [0.89, 0.96] & 0.90 [0.86, 0.95] & \textbf{0.90 [0.83, 0.96]} \\
Rich & PFE, cosine & \textbf{0.84 [0.76, 0.91]} & \textbf{1.00 [1.00, 1.00]} & \textbf{0.94 [0.91, 0.97]} & \textbf{0.91 [0.87, 0.96]} & 0.89 [0.83, 0.95] \\
Rich & PFE, MLS & 0.79 [0.70, 0.88] & 0.96 [0.92, 0.99] & 0.90 [0.86, 0.94] & 0.87 [0.82, 0.92] & 0.85 [0.77, 0.93] \\
\bottomrule
\end{tabular}%
}
\caption{Performance of the full re-identification system under five choices of step 3: relevance weighting, QAN pooling with the learned quality, QAN pooling with equal quality, and PFE fusion compared by cosine similarity or by the MLS. Steps 4 to 7 are unchanged and use the step-3 scores. Brackets give 95\% bootstrap intervals over test individuals. AUOSCR is that of step 6, because the MAP rule of step 7 does not use a fixed threshold on the known vs unknown probability.}
\label{tab:supp_late_steps}
\end{table}

\end{document}